\documentclass[conference]{IEEEtran}

\usepackage{cite}
\usepackage{amsmath,amssymb,amsfonts}
\usepackage{graphicx}
\usepackage{xcolor}
\usepackage{booktabs}
\usepackage{url}
\usepackage{enumitem}

\graphicspath{{figures/}}

\begin{document}

\title{Mobile Traffic Camera Calibration from Road Geometry for UAV-Based Traffic Surveillance}

\author{\IEEEauthorblockN{Alexey Popov, Natalia Trukhina, and Vadim Vashkelis}
\IEEEauthorblockA{Embedded Intelligence Lab}}

\maketitle

\begin{abstract}
Unmanned aerial vehicles (UAVs) can provide flexible traffic surveillance in locations where fixed roadside cameras are unavailable, costly, or impractical to install. However, raw UAV video is difficult to use for traffic analytics because vehicle motion is observed in perspective image coordinates rather than in a stable metric road coordinate system. This paper presents a lightweight pipeline for converting monocular oblique UAV traffic video into a local metric bird's-eye-view (BEV) representation. The method uses visible road geometry, such as lane markings, road borders, and crosswalks, to estimate a road-plane homography from image coordinates to metric ground-plane coordinates. Vehicle observations, supplied by dataset annotations or detectors, are projected to BEV using estimated ground contact points. The resulting metric trajectories are used to estimate vehicle direction, speed, heading, and dynamic 3D cuboids on a road plane. We evaluate the pipeline on the UAVDT dataset using ground-truth vehicle annotations to isolate calibration and geometric reconstruction performance from detector and tracker errors. We present detailed results for the M1401 sequence, in which 40 sampled frames from the source span img000001--img000196 produce 632 metric cuboid instances across 23 tracks. The experiments show that road-geometry calibration can transform monocular UAV footage into interpretable traffic-camera-style analytics, including BEV tracks and synchronized 3D cuboid visualizations. The results also reveal key limitations: far-field vehicles are sensitive to small homography errors, manual validation is currently more reliable than fully automatic calibration, and the single-plane assumption constrains performance on non-planar or visually ambiguous road regions. The proposed pipeline demonstrates a practical foundation for deployable UAV traffic cameras and future real-time traffic digital-twin systems.
\end{abstract}


\section{Introduction}
Traffic monitoring systems are commonly built around fixed infrastructure: roadside cameras, gantries, radar sensors, loop detectors, or permanently installed surveillance systems. These systems can provide stable observations, but they are expensive to deploy, tied to fixed locations, and unavailable during many temporary scenarios such as construction zones, road incidents, public events, emergency traffic control, and short-term urban planning studies.

UAVs offer a complementary sensing modality. A drone can be deployed quickly, observe a road segment from an elevated viewpoint, and capture vehicle movement over multiple lanes and intersections. UAV-based monitoring is attractive because the camera position is flexible, the deployment is temporary, and the field of view can cover a large road area. However, raw UAV video is not directly equivalent to fixed-camera traffic data. Vehicle detections are measured in image pixels and are affected by perspective, altitude, viewing angle, camera motion, and road geometry. To use UAV footage for traffic analytics, the video must be mapped into a stable road-plane coordinate system.

This paper studies the following question: can an oblique monocular UAV traffic video be converted into a metric road-plane representation suitable for traffic surveillance using visible road geometry? We propose a practical calibration and visualization pipeline that treats a UAV as a deployable mobile traffic camera. Instead of requiring pre-installed camera calibration, the method estimates a local image-to-ground homography from road-plane cues. Once calibrated, vehicle bounding boxes are projected into a metric BEV coordinate system, producing trajectories, headings, approximate speeds, and dynamic 3D cuboids.

The objective is not high-fidelity photorealistic reconstruction of vehicle meshes. Moving vehicles in traffic footage are poorly suited for static-scene reconstruction methods such as structure-from-motion, NeRF, or 3D Gaussian Splatting because they violate the static-scene assumption and are often visible for only a few consecutive frames. Instead, we target a fast and interpretable traffic-surveillance representation: object position, direction, speed, trajectory, and occupancy on the road plane.

The contributions of this paper are:
\begin{itemize}[leftmargin=*]
    \item a lightweight UAV traffic-camera calibration pipeline that maps monocular UAV video into a local metric BEV coordinate system using road geometry;
    \item a human-in-the-loop road-plane calibration workflow that supports visual validation and correction of homography parameters using lane markings and road-plane points;
    \item a dynamic traffic-scene representation that converts 2D vehicle annotations into metric BEV trajectories and 3D cuboids;
    \item a case study on UAVDT M1401, showing that UAV traffic footage can be converted into synchronized original-video, BEV, and 3D cuboid visualizations while identifying calibration sensitivities and failure modes.
\end{itemize}

The source code for reproducing the results presented in this paper is publicly available at \url{https://gitlab.com/emilab-group/uav-bev-3d}.

\section{Related Work}
\subsection{UAV-Based Traffic Monitoring}
UAVs have increasingly been studied as flexible traffic sensors because they can observe road segments from above without requiring fixed infrastructure. UAV traffic monitoring pipelines commonly include vehicle detection, tracking, scale estimation, and speed estimation. Byun et al. proposed a UAV-based road traffic monitoring method using deep neural networks for vehicle detection and tracking, lane-distance-based scale calculation, and speed estimation from aerial video \cite{byun2021road}. Tilon et al. proposed UAV vehicle tracking and speed estimation designed for deployment on embedded UAV hardware \cite{tilon2023vehicle}. These works motivate the use of UAVs as traffic sensors but also highlight the importance of converting image displacement into ground-plane motion.

The UAVDT benchmark was introduced to support object detection, single-object tracking, and multiple-object tracking in UAV video under challenging conditions such as camera motion, high density, small object size, occlusion, weather variation, and scale variation \cite{du2018uavdt}. UAVDT contains roughly 80,000 representative frames selected from 10 hours of raw UAV videos and is annotated for detection and tracking tasks. In this work, UAVDT is used not to benchmark a detector, but as a representative monocular UAV traffic-surveillance dataset for road-plane calibration and BEV trajectory recovery.

\subsection{BEV Projection and Homography}
A common way to obtain a top-down traffic representation from a monocular camera is to estimate a planar homography between the image and the road plane. Homography-based inverse perspective mapping converts road-plane pixels into a BEV image when the observed region is approximately planar. The image-to-ground transformation can be represented as
\begin{equation}
    \lambda \begin{bmatrix} x \\ y \\ 1 \end{bmatrix}
    = \mathbf{H}_{I \rightarrow G}
    \begin{bmatrix} u \\ v \\ 1 \end{bmatrix},
\end{equation}
where $(u,v)$ are image coordinates, $(x,y)$ are ground-plane coordinates, $\lambda$ is a projective scale, and $\mathbf{H}_{I \rightarrow G}$ is a $3\times3$ homography.

Geometry-based BEV approaches are computationally efficient and interpretable, but they depend strongly on calibration quality. Small errors in road-plane point selection or metric scale can create large errors in far-field vehicle positions. This sensitivity is especially important for oblique UAV footage, where far-away vehicles occupy few pixels and small image-coordinate errors translate into large ground-plane errors.

\subsection{Camera Calibration for Traffic Surveillance}
Traffic camera calibration aims to recover the relationship between image coordinates and real-world road coordinates. For fixed roadside cameras, calibration can exploit lane markings, road layout, vehicle motion, or known scene geometry. Recent work has explored automated traffic-camera calibration using learned models and road topology. D'Amicantonio et al. proposed automated camera calibration via homography estimation with graph neural networks for traffic-camera scenes \cite{damicantonio2024automated}. Our setting differs from fixed roadside calibration in two ways: the camera is mounted on a UAV and may be temporary or mobile, and the calibration workflow is designed for deployable surveillance where pre-installed calibration targets or surveyed camera parameters may not exist.

\subsection{BEV Representations in Autonomous Driving}
BEV representations are also central to autonomous-driving perception. Camera-based BEV methods such as Lift-Splat-Shoot lift image features into 3D frustums and splat them into a BEV grid for downstream planning and perception \cite{philion2020lift}. These systems typically estimate task-oriented object states rather than reconstructing photorealistic meshes. Our work adopts a similarly task-oriented representation for UAV traffic surveillance: a dynamic metric scene composed of a road plane and vehicle cuboids.

\section{Problem Statement}
Given an oblique monocular UAV video of a road segment and vehicle observations in image coordinates, the goal is to recover a local metric traffic-scene representation.

\subsection{Inputs}
The pipeline assumes the following inputs:
\begin{itemize}[leftmargin=*]
    \item UAV video frames;
    \item vehicle detections or ground-truth annotations;
    \item road-plane calibration cues, such as lane markings, road borders, crosswalks, or manually selected road-plane points;
    \item optional class labels such as car, truck, bus, or generic vehicle.
\end{itemize}

\subsection{Outputs}
The desired outputs are:
\begin{itemize}[leftmargin=*]
    \item a metric BEV coordinate system for the observed road region;
    \item vehicle trajectories in ground-plane meters;
    \item vehicle heading and approximate speed;
    \item dynamic 3D cuboids placed on the road plane;
    \item synchronized visualizations: original image, BEV tracks, and 3D cuboid scene.
\end{itemize}

\subsection{Assumptions}
The current method assumes that the relevant traffic surface is locally planar, vehicle ground contact points can be approximated from 2D bounding boxes, road geometry is visible enough to calibrate a homography, vehicle annotations or detections are available, and the UAV camera is sufficiently stable over the analyzed segment or can be treated with a single local calibration.

\section{Method}
\subsection{Overview}
The pipeline converts UAV image observations into a dynamic metric traffic scene:
\begin{equation}
    \text{frames} \rightarrow \text{boxes} \rightarrow \text{homography} \rightarrow \text{BEV tracks} \rightarrow \text{cuboids}.
\end{equation}
Unlike dense 3D reconstruction, the pipeline produces a traffic-analytics representation that is fast, interpretable, and robust enough for short UAV surveillance sequences.

\subsection{Frame Sampling and Mapping}
A short subsequence is sampled from the UAV video. A frame mapping table preserves the relationship between sampled frame names and original UAVDT frame indices. This mapping is essential because annotation misalignment can create false trajectory motion. During development, using an incorrect frame step caused annotation boxes to move approximately twice as fast as the visible vehicles. Therefore, every sampled frame is associated with its original frame index and the annotation table is joined through this mapping rather than through an assumed fixed step alone.

\subsection{Vehicle Observations from Annotations}
For experiments, UAVDT ground-truth annotations are used instead of detector outputs. This isolates the calibration and reconstruction problem from detection and tracking errors. For each object annotation we use
\begin{equation}
    (f, id, x_0, y_0, w, h, c),
\end{equation}
where $f$ is the original frame index, $id$ is the target identity, $(x_0,y_0,w,h)$ is the bounding box, and $c$ is the object category. The box corners are
\begin{equation}
    x_1=x_0, \quad y_1=y_0, \quad x_2=x_0+w, \quad y_2=y_0+h.
\end{equation}
The annotation target identity is used directly as the track identity. This avoids identity switches observed with a simple nearest-neighbor tracker.

\subsection{Road-Plane Calibration}
The core of the method is the image-to-ground homography $\mathbf{H}_{I\rightarrow G}$. Calibration is obtained from road-plane point correspondences:
\begin{equation}
    (u_i,v_i) \leftrightarrow (x_i,y_i), \quad i=1,\ldots,N,
\end{equation}
where $(u_i,v_i)$ are image points and $(x_i,y_i)$ are metric ground-plane points. The homography is estimated with more than four correspondences when available, using lane markings, road borders, crosswalk corners, or manually selected road-plane points.

The workflow supports human-in-the-loop calibration. The operator selects visible road-plane points and assigns metric coordinates based on lane width and road direction. Calibration is validated by: (i) warping the road image into metric BEV, (ii) drawing a metric grid in BEV, and (iii) reprojecting the metric grid back onto the original image. A mathematically valid homography can still be geometrically poor if selected points are not on the road plane or if assigned metric coordinates are inconsistent. Visual validation is therefore a central part of the method.

\begin{figure}[t]
    \centering
    \includegraphics[width=\linewidth]{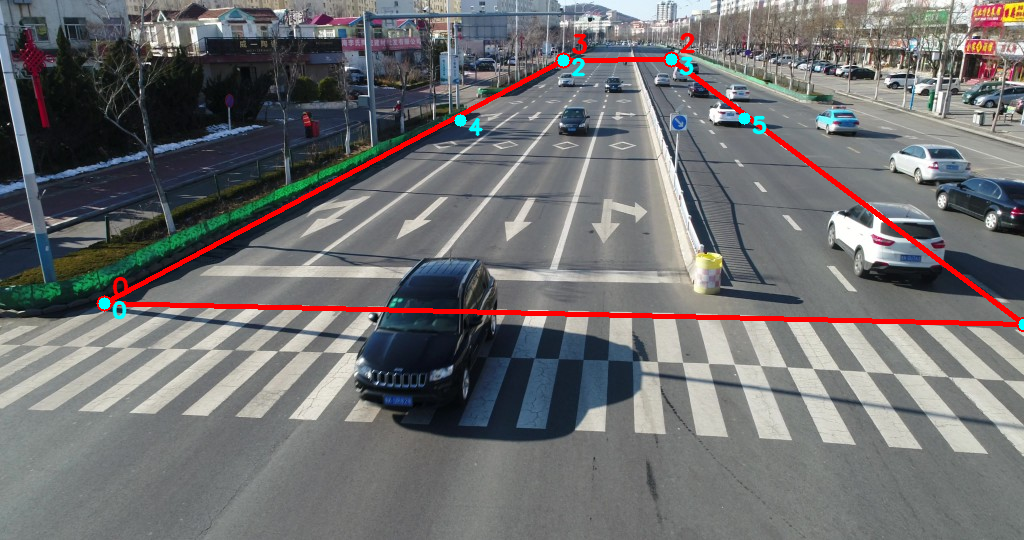}
    \caption{Road-geometry calibration on the UAVDT M1401 reference frame. Red points define the active road-plane correspondences used for the metric homography; cyan points provide additional visual checks on lane and road geometry. The calibrated ground plane spans approximately 24 m across the road and 90 m along the road.}
    \label{fig:calibration_image}
\end{figure}

\subsection{Vehicle Ground-Point Projection}
For each vehicle bounding box, a ground contact point is approximated. The default choice is the bottom center:
\begin{equation}
    u_g = \frac{x_1+x_2}{2}, \qquad v_g = y_2.
\end{equation}
This point is projected to the metric road plane:
\begin{equation}
    \lambda \begin{bmatrix} x_g \\ y_g \\ 1 \end{bmatrix}
    = \mathbf{H}_{I\rightarrow G}
    \begin{bmatrix} u_g \\ v_g \\ 1 \end{bmatrix}.
\end{equation}
The bottom-center point is an approximation. In oblique UAV views, the detector box may include roof pixels, shadow, or partial object extent, so footprint-aware or segmentation-based ground-point estimation may improve accuracy.

\subsection{Trajectory, Speed, and Heading}
For each track identity, ground positions are sorted by frame. Velocity is estimated by finite differences:
\begin{equation}
    v_x = \frac{x_{t+1}-x_{t-1}}{t_{t+1}-t_{t-1}}, \quad
    v_y = \frac{y_{t+1}-y_{t-1}}{t_{t+1}-t_{t-1}}.
\end{equation}
The speed and heading are
\begin{equation}
    s = \sqrt{v_x^2+v_y^2}, \qquad
    \theta = \operatorname{atan2}(v_y,v_x).
\end{equation}
On straight roads, headings may be regularized toward a dominant road axis to reduce jitter from annotations and projection noise.

\subsection{Vehicle Cuboid Generation}
Each object is represented as an oriented metric cuboid:
\begin{equation}
    \mathbf{c}=\left[x_g, y_g, \frac{h_c}{2}\right]^T,
\end{equation}
with dimensions $(l_c,w_c,h_c)$ and yaw $\theta$. Class-dependent dimension priors are used, as shown in Table~\ref{tab:dims}.

\begin{table}[t]
\caption{Vehicle dimension priors used for cuboid generation.}
\label{tab:dims}
\centering
\begin{tabular}{lccc}
\toprule
Class & Length (m) & Width (m) & Height (m) \\
\midrule
Car & 4.5 & 1.8 & 1.5 \\
Truck & 8.0 & 2.5 & 3.0 \\
Bus & 12.0 & 2.6 & 3.2 \\
Generic vehicle & 4.5 & 1.8 & 1.5 \\
\bottomrule
\end{tabular}
\end{table}

\subsection{Visualization and Validation}
The pipeline generates synchronized outputs: original UAV frames with annotations, metric BEV trajectories, dynamic 3D cuboid scenes, side-by-side validation videos, and interactive 3D scene views. These visualizations are not only presentation artifacts but also diagnostics. Misalignment between original boxes, BEV points, and cuboids reveals errors in frame mapping, homography calibration, ground-point projection, or cuboid generation.

\begin{figure}[t]
    \centering
    \includegraphics[width=0.78\linewidth]{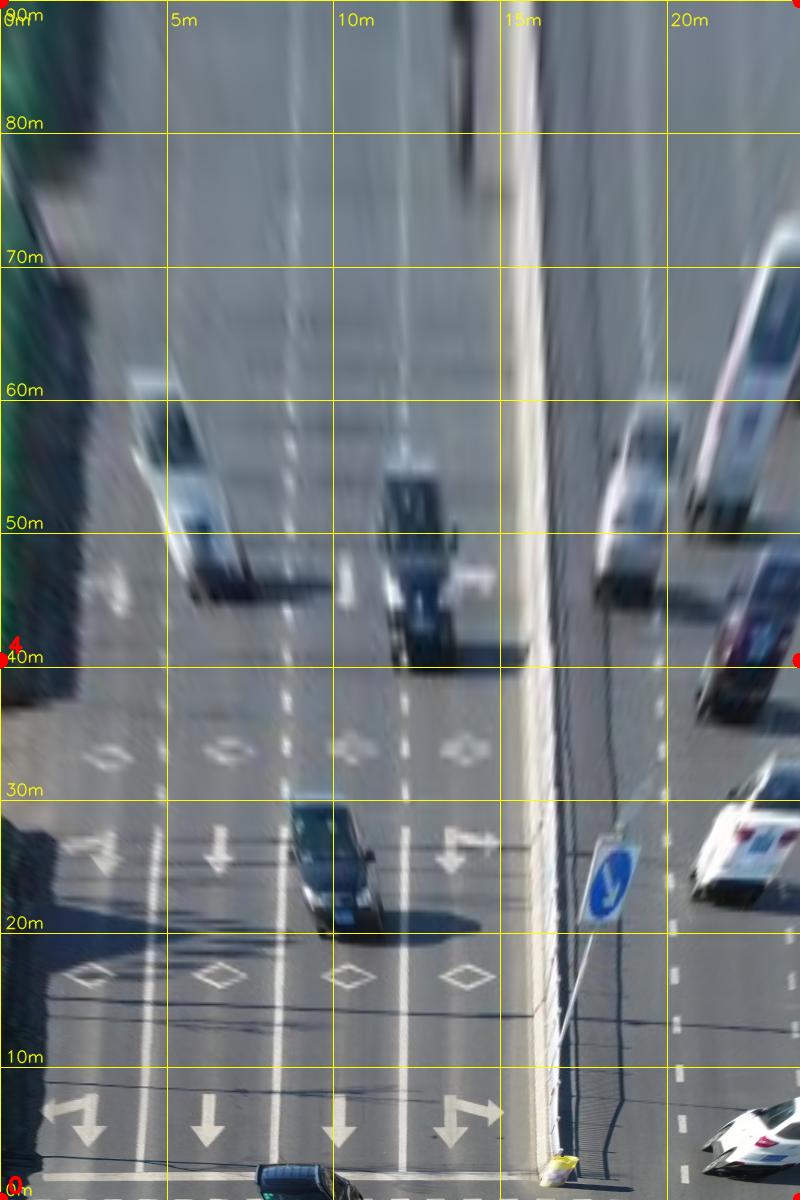}
    \caption{Metric BEV preview produced by the calibrated homography. The grid provides a direct visual check that the road direction, lateral scale, and metric extents remain plausible after inverse perspective mapping.}
    \label{fig:metric_bev_grid}
\end{figure}

\section{Experiment}
\subsection{Dataset and Sequence}
We evaluate the pipeline on the UAVDT dataset and present detailed results for sequence M1401, an oblique UAV traffic-surveillance sequence showing a multi-lane urban road with visible lane markings, crosswalks, and moving vehicles. UAVDT contains annotated UAV videos for object detection and tracking, making it suitable for controlled experiments on traffic-scene projection and calibration \cite{du2018uavdt}.

\subsection{Experimental Setup}
Table~\ref{tab:setup} summarizes the current experimental configuration.

\begin{table}[t]
\caption{Dataset and processing settings for the current run.}
\label{tab:setup}
\centering
\begin{tabular}{ll}
\toprule
Item & Value \\
\midrule
Dataset & UAVDT \\
Sequence & M1401 \\
Source span & img000001--img000196 \\
Sampling interval & 5 frames \\
Rendered frames & 40 \\
Vehicle observations & Ground-truth annotations \\
Calibration & Human-in-the-loop metric homography \\
Road-plane size & 24 m $\times$ 90 m \\
Output & BEV tracks and dynamic 3D cuboids \\
\bottomrule
\end{tabular}
\end{table}

\subsection{Experimental Variants}
We consider four variants in the development and validation process: automatic line-based homography using line detection and vanishing-point heuristics; manual road-corner calibration using four or more road-plane points; multi-point lane calibration using lane markings; and detector-generated observations compared with UAVDT ground-truth annotations. The GT-based path is used for the reported run in order to evaluate calibration and projection behavior without detector/tracker confounding.

\section{Results}
\subsection{Qualitative Results}
The calibrated UAV sequence can be transformed into a coherent metric BEV representation. Vehicle annotations project onto the road plane, forming trajectories consistent with traffic flow. Dynamic cuboids provide an interpretable 3D traffic scene in which vehicle positions, directions, and approximate scale are visible.

The current run produces synchronized original-video, BEV-track, and 3D-cuboid visualizations. The original view confirms annotation alignment, the BEV view shows road-plane trajectories, and the 3D view shows class-dependent cuboids moving over the road plane. This triptych representation improves interpretability relative to raw UAV video alone because it separates image appearance from metric traffic state.

\begin{figure*}[t]
    \centering
    \includegraphics[width=\textwidth]{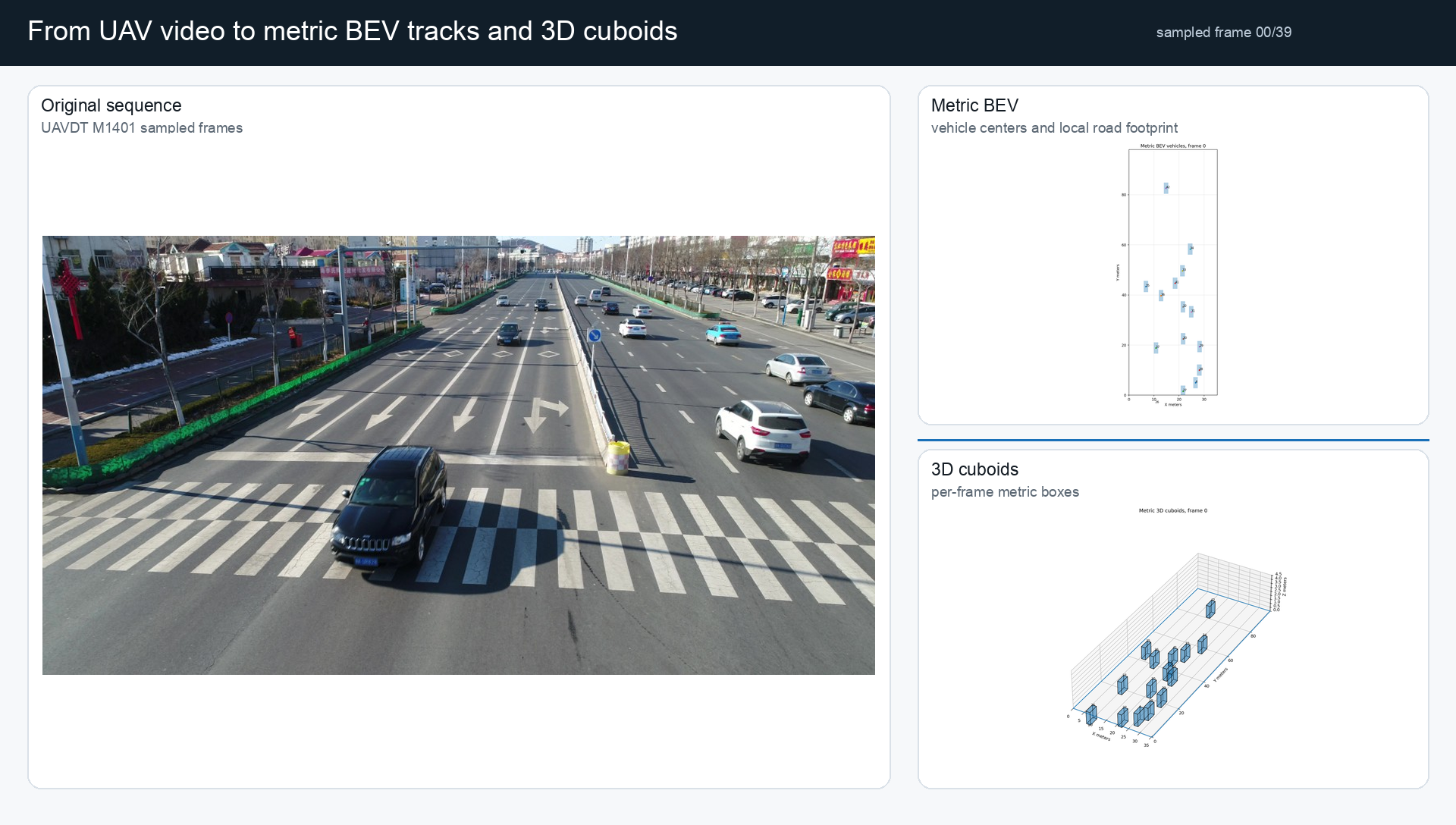}
    \caption{Synchronized demonstration frame generated by the pipeline. The left panel shows the original UAV frame, the upper-right panel shows the metric BEV view, and the lower-right panel shows the metric 3D cuboid scene. This representation illustrates the intended use of the UAV as a mobile traffic camera: raw oblique video is converted into a structured road-plane traffic state.}
    \label{fig:triptych}
\end{figure*}

\begin{figure}[t]
    \centering
    \includegraphics[width=0.58\linewidth]{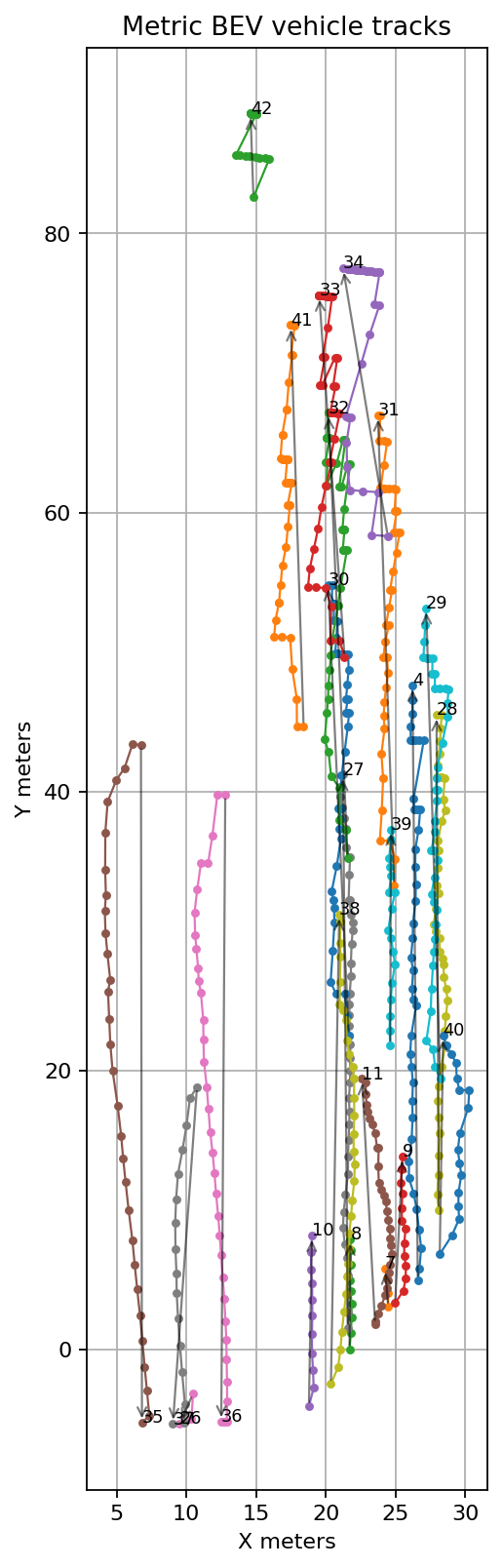}
    \caption{Metric BEV vehicle tracks for the reported M1401 run. The trajectories are expressed in the calibrated road-plane coordinate system rather than in perspective image pixels.}
    \label{fig:bev_tracks}
\end{figure}

\begin{figure}[t]
    \centering
    \includegraphics[width=\linewidth]{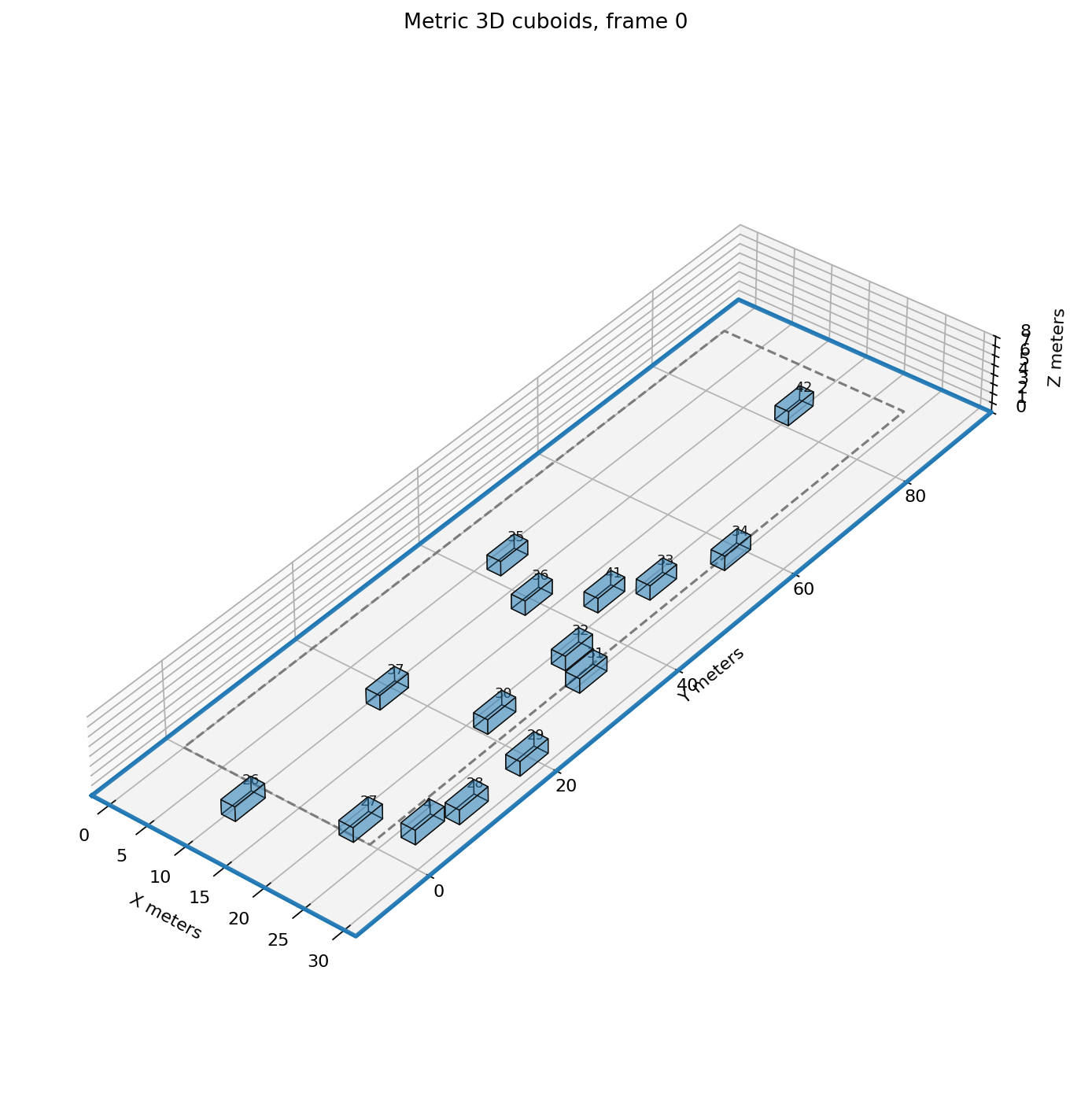}
    \caption{Metric 3D cuboid preview for one sampled frame. Cuboids use class-level dimension priors and road-axis-aware heading estimates. The visualization is intended for traffic-state interpretation rather than photorealistic 3D reconstruction.}
    \label{fig:cuboids}
\end{figure}

\subsection{Quantitative Summary}
Table~\ref{tab:results} reports summary statistics for the current run.

\begin{table}[t]
\caption{Output statistics for the current run.}
\label{tab:results}
\centering
\begin{tabular}{lc}
\toprule
Metric & Value \\
\midrule
Rendered frames & 40 \\
Unique tracks & 23 \\
Cuboid instances & 632 \\
Mean points per track & 27.5 \\
Mean track speed & 8.1 m/s \\
Coordinate system & Local metric road plane \\
Vehicle model & Oriented cuboid \\
\bottomrule
\end{tabular}
\end{table}

\subsection{Calibration Behavior}
Initial fully automatic homography estimation produced plausible but geometrically imperfect BEV views. Vehicle cuboids tended to cluster or drift when the road quadrilateral was poorly estimated. Manual lane-geometry calibration improved interpretability by aligning road markings more closely with the metric BEV grid. This supports the hypothesis that human-in-the-loop calibration is currently more robust than fully automatic calibration for real UAV traffic footage.

In the reported calibration, four active road-plane correspondences define a 24 m by 90 m metric road rectangle on the reference frame. Two additional lane-edge points were retained as visual checks rather than as active homography constraints. This separation was useful during development because adding visually plausible but geometrically inconsistent points can reduce interpretability even when the numerical least-squares fit appears valid.

\subsection{Heading and Cuboid Orientation}
Raw motion-based heading can be noisy due to annotation jitter, frame sampling, and homography sensitivity. Road-axis-aware heading estimation produces more plausible cuboid orientation, especially on straight road segments. This improves the visual consistency of the dynamic 3D scene.

\subsection{Far-Field Sensitivity}
Far-field vehicles show greater sensitivity to small homography errors than near-field vehicles. This is expected because distant vehicles occupy fewer pixels and are close to the vanishing region of the perspective projection. Small image-coordinate errors or lane-width errors can produce large lateral shifts in metric coordinates. Far-field drift therefore remains a key limitation and a useful diagnostic for calibration quality.

\section{Discussion}
\subsection{Practical Value}
The proposed system demonstrates that UAVs can function as mobile traffic cameras. Instead of requiring permanent installation, a drone can be deployed temporarily, calibrated from visible road structure, and used to recover BEV traffic trajectories. This is useful for temporary traffic studies, construction-zone monitoring, incident response, event traffic management, emergency routing, and short-term road safety analysis.

In practice, deployed UAV surveillance systems must transmit video over bandwidth-limited wireless links, which can degrade the fine local detail needed for reliable detection and calibration. Trukhina and Vashkelis~\cite{trukhina2026hybrid} proposed a hybrid visual telemetry scheme in which a continuous low-bitrate HEVC video stream is augmented by selectively transmitted high-detail JPEG stills of regions of interest, enabling both scene-level awareness and close-up object analytics under constrained communication budgets. Such bandwidth-aware transmission is complementary to the proposed pipeline and important for end-to-end deployable UAV traffic cameras.

\subsection{Why Not Full 3D Reconstruction?}
Moving vehicles are difficult for classical 3D reconstruction or 3D Gaussian Splatting because they violate static-scene assumptions. Each vehicle may appear in only a small number of frames and its position changes independently of the camera. For traffic analytics, detailed mesh reconstruction is unnecessary. A metric cuboid representation is more practical, faster, and better aligned with traffic-surveillance needs.

\subsection{Limitations}
The largest limitation is homography sensitivity. Calibration quality degrades when lane markings are weak or occluded, road borders are ambiguous, selected points are not on the road plane, the road is curved or non-planar, vehicles are far from calibration points, or the UAV moves significantly during the sequence. The current method also uses approximate vehicle ground-contact points and class-level cuboid dimension priors rather than measured vehicle dimensions. In the selected UAVDT sequence, all rows in the available \texttt{gt\_whole} annotation file are labeled with object category 1, which corresponds to cars in the UAVDT detection format. Therefore, the reported cuboids should be interpreted as vehicle-level car cuboids; motorcycles or bicycles visible in the raw imagery are not separately available from this annotation source.

\subsection{Future Work}
Future work should address automatic lane-level calibration, multi-frame homography stabilization, road-axis and lane-centerline extraction, robust ground-contact point estimation, integration with modern detectors and trackers, real-time processing, multi-drone fusion, georeferenced coordinates using GPS/IMU, curved-road and multi-plane road modeling, and uncertainty estimation for far-field vehicles.

\section{Conclusion}
This paper presented a lightweight pipeline for converting monocular UAV traffic video into a metric road-plane representation. By calibrating the UAV view from visible road geometry, vehicle annotations can be projected into BEV coordinates, producing trajectories, headings, approximate speeds, and dynamic 3D cuboids. Experiments on the UAVDT dataset, with detailed results for sequence M1401, show that drones can serve as deployable mobile traffic cameras when their video is calibrated into a local metric coordinate system. The results demonstrate practical traffic-scene reconstruction from monocular UAV footage while highlighting the importance of calibration validation, especially for far-field vehicles. The proposed system provides a foundation for mobile UAV-based traffic analytics and future traffic digital-twin applications.

\end{document}